\documentclass[conference]{IEEEtran}

\IEEEoverridecommandlockouts
\usepackage{cite}
\usepackage{amsmath,amssymb,amsfonts}
\usepackage{algorithmic}
\usepackage{graphicx}
\usepackage{textcomp}
\usepackage{subfigure}
\usepackage{xcolor}
\usepackage{multirow}

\def\BibTeX{{\rm B\kern-.05em{\sc i\kern-.025em b}\kern-.08em
    T\kern-.1667em\lower.7ex\hbox{E}\kern-.125emX}}
\begin{document}

\title{Physics-Informed Graph Learning\\
}

\author{\IEEEauthorblockN{Ciyuan Peng}
\IEEEauthorblockA{\textit{ Institute of Innovation, Science and Sustainability} \\
\textit{Federation University Australia}\\
Ballarat, Australia \\
ciyuan.p@outlook.com}
\\
\IEEEauthorblockN{Vidya Saikrishna}
\IEEEauthorblockA{\textit{Global Professional School} \\
	\textit{Federation University Australia}\\
	Ballarat, Australia \\
	v.saikrishna@federation.edu.au} 
\and
\IEEEauthorblockN{Feng Xia$^{\ast}$ \thanks{*Corresponding Author}}
\IEEEauthorblockA{\textit{Institute of Innovation, Science and Sustainability} \\
	\textit{Federation University Australia}\\
	Ballarat, Australia \\
	f.xia@ieee.org}

%\and

\\
\IEEEauthorblockN{Huan Liu}
\IEEEauthorblockA{\textit{School of Computing and Augmented Intelligence} \\
\textit{Arizona State University}\\
Tempe, AZ, USA \\
huan.liu@asu.edu}

}

\maketitle

\begin{abstract}
An expeditious development of graph learning in recent years has found innumerable applications in several diversified fields. Of the main associated challenges are the volume and complexity of graph data. The graph learning models suffer from the inability to efficiently learn graph information. In order to indemnify this inefficacy, physics-informed graph learning (PIGL) is emerging. PIGL incorporates physics rules while performing graph learning, which has enormous benefits. This paper presents a systematic review of PIGL methods. We begin with introducing a unified framework of graph learning models followed by examining existing PIGL methods in relation to the unified framework. We also discuss several future challenges for PIGL. This survey paper is expected to stimulate innovative research and development activities pertaining to PIGL.
\end{abstract}


\section{Introduction}

Graph learning (GL) is a rapidly growing artificial intelligence (AI) technique that refers to applying machine learning on graph data \cite{xia2021graph}. It has attracted great attention in recent years. Many GL algorithms, such as graph neural networks (GNNs) and random walks \cite{Xia2019TETCIrandom}, have shown  great capability to learn complex relations and dependencies between vertices of graphs, and capture rich knowledge embedded in graphs \cite{wang2021graph}. This makes, GL, a primary approach for graph analysis and also effective for accomplishing various downstream tasks. In the subsections below, we first discuss the challenges facing graph learning, followed by some solutions for addressing the challenges and at the end, we outline the contributions of this survey paper.

\subsection{The Challenges for Graph Learning}
Despite the popularity of  GL algorithms, a lot of challenges still remain for GL. We discuss some major challenges below.

The first challenge is concerned with the success of data-driven models. The performance of data-driven models is highly dependent on the quality and availability of data. In view of this issue, data processing has become a crucial step in GL methods. 
The biggest challenge is to efficiently extract valid knowledge and information from data deluge \cite{chao2022fusing}. Owing to the complex nature of graph data, it is difficult to collect data that has no quality issues~\cite{3524718}. Few examples of quality problems include: false information (e.g., the wrong triplet in knowledge graphs), missing data (e.g., missing important nodes or edges) and data uncertainty (e.g., corrupted or distorted data). Therefore, imperfection in graph data often leads to the degradation of performance in GL models. 

The second challenge concerns the embedding of graph data into feature space in which the model tries to learn hidden characteristics of data. The learning method requires to keep intact the structural information of data~\cite{wang2020exploiting}. However, many existing GL algorithms cannot fully preserve the structure of graph data~\cite{sun2020graph}. An example is random negative sampling of nodes  \cite{grover2016node2vec}, because of which the features of graph structure can easily get destroyed. There is a consequence in the destruction of data structure during graph representation.

Thirdly, most existing GL models initialize parameters randomly and heavily rely on manual parameters optimization to achieve better performance. This could result in costly human efforts and low efficiency of models. 

Finally, many GL systems are black boxes or unintelligible to human beings~\cite{henderson2021improving}. 
%This is especially true for GNNs due to the black-box nature of neural networks. 
Deep learning models generally conduct training without any support of scientific principles. To achieve trust and transparency, there is a strong need for explainability or interpretability of the models.

\subsection{Overcoming Challenges with Physics-Informed Graph Learning}

Inspired by physical-informed machine learning algorithms, some researchers have worked on integrating physics principles with GL calling it as physics-informed graph learning (PIGL). PIGL refers to the graph learning models are trained from the additional information gained by enforcing physical laws, such as the distribution rule of continuous space. PIGL solves various challenges faced by GL\cite{9839583,jia2021physics,li2021physics}.

As traditional GL models are incapable of dealing with a variety of problems caused by complex data, physics-based graph data processing approaches have emerged. 
When dealing with imperfect data in data processing step, PIGL models can combine physical information with data to transform useless data into valid data~\cite{da2020combining,xiang2021physics}. Specially, PIGL intergrates flawed data and mathematical models reflecting physical laws, and then implements them through graph neural networks.
For example, Yang et al. \cite{yang2021b} integrated the Bayesian model with physics-informed neural networks for quantifying the uncertainty of noisy data. In addition, by embedding and enforcing the physics principles, physical models can generate simulation data to enrich sparse data and even pre-train the GL models to reduce the required amount of data.

In the graph representation process, the structure of graph data is easily destroyed in the feature space because the relational structure is ignored~\cite{3398518,3367303}. When the layout of network structure is influenced by physical principles (e.g., repulsive and attractive force), the graph structure can be preserved in the feature space by utilizing physics constraints to model the interaction of nodes \cite{haleem2019evaluating,Yu2019TCSSacademic}.

It is worth noting that some PIGL models can reasonably initialize model parameters to improve the efficiency of the model and reduce the model overhead. The core idea is to physically inform the initial states of the learning models that satisfy some physics principles \cite{willard2020integrating}. The main method of physics-based initialization revolves around using synthetic data generated by physical approaches that are used to pre-train the GL models \cite{almajid2022prediction}. In this way, not only the model parameters can be reasonably initialized, the data paucity issues can also be alleviated. Moreover, physics principles can provide scientific theories to elucidate the inner mechanisms of deep learning \cite{geiger2020scaling}.

\subsection{Contributions}
In this survey, we provide a systematic overview of PIGL methods. To the best of our knowledge, this is the first survey paper on this emerging field. We first discuss the unified framework of GL and explain the three main steps required to implement GL. Then, this paper elucidates the significance of PIGL models from three aspects:  physics-based graph data processing, graph representation with physical properties, and physics-driven learning models from the perspective of the unified framework of GL. Following this, the future directions of PIGL are outlined. The last section finally concludes the paper.

\section{A Unified Framework of Graph Learning}
\begin{figure*}
	\centering
	\includegraphics[width=5.5in]{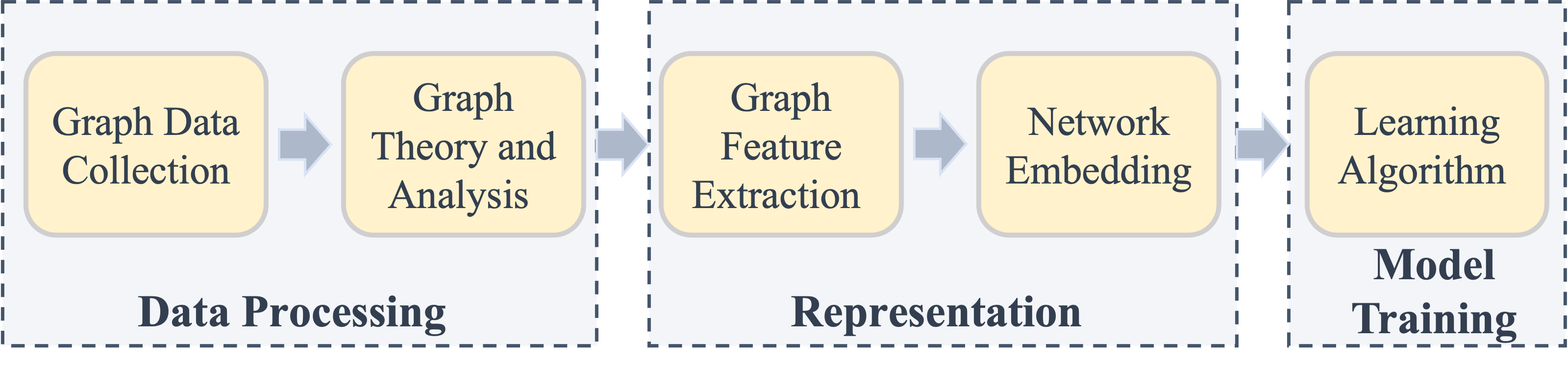}
	\caption{The unified framework of graph learning.}
	\label{1}
\end{figure*}

This section discusses the unified framework of GL (see Fig.~\ref{1}). The  three main steps to be considered for  graph learning are: data processing, representation, and model training.

\subsection{Data Processing}\label{Data Processing}
The data processing step first requires collection of corresponding graph data based on the needs of tasks and applications. This is normally followed by data analysis for future processing. 

\subsubsection{Graph Data Collection}
Unlike unstructured data, graph data represents not only the vertices, but also the relationships between vertices are included. It is, therefore, more complex than unstructured data. A vast number of graph databases, such as TigerGraph\footnote{https://www.tigergraph.com/} and Neo4j\footnote{https://neo4j.com/}, have been made public for various projects. Also, many benchmark graph datasets are built for a variety of research. For example, Cora dataset \cite{cabanes2013cora}, a citation network, can be used for academic network analysis.

\subsubsection{Graph Theory and Analysis}
A graph is represented as $G = (V, E)$, where $V$ indicates the set of nodes (vertices), and $E$ is the set of edges. $e_{ij}$ is the edge between node $v_i$ and $v_j$, denoted as  $e_{ij}=(v_i,v_j)\in E$. Graphs are categorosed as: undirected and directed graphs \cite{wu2020comprehensive}. Directed graphs also known as digraphs are defined as graphs in which edges have direction. The directions are usually indicated by arrow on the edges. Undirected graphs are special case of directed graphs where direction is not mentioned or on other words the edges are bi-directional. 

\subsection{Representation}

The second step of GL is graph representation. This step requires extraction of  graph features first and then it is followed by network embedding.

\subsubsection{Graph Feature Extraction}

To facilitate graph embedding algorithms,  the graph features are first obtained. Among the features of graph, extraction of node features (e.g., research topics of scholars in citation networks) is especially important~\cite{Sun2020IACCESS,liu2021TKDE,Xia2019TITSranking}.  For a set of $n$ nodes ${\{v_1,v_2,...,v_n}\}$ with $d$ features, the feature vector of a node $v$ can be indicated as $x_v\in \mathbb{R}^d$, and the feature matrix of the graph is $X\in \mathbb{R}^{n\times d}$. In general, the features of graphs are high dimensional. Therefore, dimensionality reduction, projecting the high-dimensional features into a new low-dimensional space, is needed~\cite{lu2021low}. 

\subsubsection{Network Embedding}	
Network embedding (a.k.a graph embedding) aims at finding a mapping function to translate nodes in the original network into low-dimensional and dense vector forms~\cite{xue2022dynamic,Hou2020CSR}. This helps in the node vectors to be easily utilised by  the machine learning models. In the vector space, the relations among nodes are represented by the distance between node vectors. Some  of well-known widely used network embedding models are random walk, matrix factorization, and deep neural networks \cite{cui2018survey}.

\begin{table*}
	\centering
	\caption{Some studies of PIGL.}	
	\newcommand{\tabincell}[2]{\begin{tabular}{@{}#1@{}}#2\end{tabular}}	
	
	\begin{tabular}{|p{0.4\columnwidth}|l|l|p{0.7\columnwidth}|}
		\hline
		Type &Objective& Reference& Approach\\
		\hline
		\multirow{3}{*}{\tabincell{l}{Physics-based Graph \\Data Processing} }&Sparse data enhancement& \cite{li2021physics} & Utilizing physical sparsity property of data to locate unlabelled data. \\ 
		\cline{2-4}
		&Filling missing data&\cite{seo2019differentiable}& Incorporating differentiable physics equations with graph learning.\\
		\cline{2-4}
		&Noisy value removal& \cite{salehi2021physgnn} & Incorporating the physical characteristics with GNNs to capture the noisy values.\\
		\hline
		\multirow{2}{*}{\tabincell{l}{Graph Representation\\ with Physical Properties}}&Precise graph visualisation&\cite{haleem2019evaluating}& Combining force-directed graph layout algorithm and deep learning algorithms.\\
		\cline{2-4}
		&Graph data structure preserving& \cite{sun2020graph}  & Embedding the principles of force interaction into graph learning models.     \\
		\hline
		\multirow{3}{*}{\tabincell{l}{Physics-driven Learning \\Models } } &Data dependency reduction& \cite{shah2018airsim} & Generating simulated data to pre-train the driving algorithm.\\
		\cline{2-4}
		&Reasonable parameter initialization& \cite{jia2021physics} & Generating simulated physical variables to pre-train recurrent graph network models.   \\
		%&Accurate and explicable learning & \cite{schuetz2021combinatorial} & Physics-driven loss function    \\
		\cline{2-4}
		&Accurate and explicable learning&\cite{daw2020physics}&Physics-driven loss function of LSTM model.\\
		\hline
	\end{tabular} 
	\label{table}
\end{table*}

\subsection{Model Training}

The final step of GL is model training, which aims to learn the features of training data and get augmented with dealing of downstream tasks. To learn the features of training data, the learning algorithm is first designed. %The learning algorithm is based on the necessity of task to be performed like classification. 
The training and validation data sets are then used to build and evaluate the model, respectively. In order to obtain a model with the high performance, the model should be tuned according to the validation results. Finally, the model is tested by the test data set.

\section{Physics-Guided Design of Graph Learning Models}
In this section, we discuss some representative existing PIGL methods, and classify these methods by the framework of GL. Firstly, we introduce some works that are related to physics-based graph data processing. Then, the physics-inspired graph representation methods are discussed. Finally, we summarize how to embed physics into the learning algorithms. Table~\ref{table} gives some representative research on PIGL.

\subsection{Physics-based Graph Data Processing}

As graph data is complex and data quality cannot always be guaranteed, the step of graph data processing is quite challenging~\cite{3524718}. Specifically, the performance of GL models could seriously be impacted by imperfect data that includes sparse, noisy and incomplete data. To alleviate this challenge, the use of physical information to enhance graph data has attracted much attention recently~\cite{iakovlev2020learning,sun2021physics}. 

Sparse data (e.g., low ratios of labelled data) is one of the most challenging issues for data collection~\cite{9709524}. Therefore, some studies suggest to combine physical models to deal with such problems. Representatively, Li and Deka \cite{li2021physics} took use of physical information contented in observation data to achieve sparse data enhancement and proposed a physics-informed graph neural network model to locate faults in power grids. The power grid resembles a graph, a fault at one point in the power grid results in abnormal data (voltage, current), and the faulty position is regarded as the label of the fault. To detect the faults in power grids using data-driven models, correctly labelled data sets are required. However, the collected power grid data sets are usually insufficiently labelled data because of the complexity and variability of the power grid. To better address this problem, the physical sparsity property of data is leveraged to reduce the requirements on labelled data. 

Consider a power grid as a graph with $n$ nodes. $\mathbf{Y}\in \mathbb{C}^{3\times 3}$ is the admittance matrix. When there is no fault, the voltages and currents of all the points in the power grid are in normal conditions, denoted as $\mathbf{U}_0\in \mathbb{C}^{3\times 1}$ and $\mathbf{C}_0\in \mathbb{C}^{3\times 1}$, respectively. The formula $\mathbf{Y}\mathbf{U}_0=\mathbf{C}_0$ can be satisfied. While if there is a fault at node $f$, which is between the nodes $i$ and $j$, the voltages and currents of the power grid will be changed. The real voltages and currents are $\mathbf{U}$ and $\mathbf{C}$, and according to the Kirchhoff’s law and the substitution theory \cite{majidi2014fault}, the following formula holds:

\begin{equation}
	\label{equ1}
	\mathbf{YU}-\Delta_{ij}=\mathbf{C}.
\end{equation}
Here, $\Delta_{ij}\in \mathbb{C}^{3n}$ is a sparse vector with nonzero values corresponding to the two nodes $i$ and $j$. Then, according to physical laws of conservation, let $\Delta \mathbf{U} = \mathbf{U} - \mathbf{U}_0$ and $\Delta \mathbf{C} = \mathbf{C} - \mathbf{C}_0$. Thereby, the Equation~\ref{equ1} can be interpreted as:

\begin{equation}
	\label{equ2}
	\mathbf{Y}\Delta \mathbf{U}=\Delta \mathbf{C} + \Delta_{ij}.
\end{equation}

The summation of $\Delta \mathbf{C}$ and $\Delta_{ij}$ is the physical interpretation of the linear combinations. Due to the non-zero values in $\Delta_{ij}$,  at nodes $i$ and $j$, the dominant entries of $\Delta \mathbf{C} + \Delta_{ij}$ are closely related to the fault location $f$.
Therefore, when the labelled faults in the observed data are lacking, this physical information of voltages and currents can help to estimate the unlabelled faults and enhance the sparse data. Especially, utilizing the physical similarity of unlabelled and labelled data to transform useless sparse data into valid data is a trending topic for graph data processing~\cite{zhang2022physics}.

Li and Deka \cite{li2021physics} implemented their model in the 123-node test feeder ~\cite{kersting1991radial}. Three performance metrics including location accuracy rate ($LAR$),  $LAR^{1-hop}$, and F1-score were adapted. Compared with three well-known baselines: neural network, convolutional neural network, and graph convolutional network, their model showed better locates various faults at low ratios of labeled data. Thereby, it shows the physics-based graph data processing can enhance the sparse data and improve the model performance.

In addition to overcoming the sparse data problems, other common problems associated with data processing (e.g., noisy values and missing data) are also expected to be solved by physical methods. For example, Seo and Liu~\cite{seo2019differentiable} presented DPGN (differentiable physics-informed graph networks) model incorporating differentiable physics equations with the graph learning. They transformed physical information in the physical model into usable data to fill the missing values. Therefore, the known physical rules are used to compensate for incomplete graph data caused by inadequate observations. 

Salehi and Giannacopoulos \cite{salehi2021physgnn} proposed PhysGNN model, which combines the physical characteristics of human soft tissue and graph neural networks, to analyse the preoperative data and guide the neurosurgical procedures. The preoperative data has the property to constantly change and be imprecise. This results in a lot of noise being present in the data set. They took advantage of the physical characteristics of human soft tissue to capture the noisy values, so that the data set can be more accurate.

\subsection{Graph Representation with Physical Properties}

In graph representation step, to preserve the structure of graph data in the feature vector space, some studies import physical constraints when embedding data into low-dimensional space. In particular, adding physical properties to the input feature vector can ensure that the graph structure is not damaged to a certain extent and the data features are better preserved in the vector space.

In the field of graph visualisation \cite{Liu2018IACCESSsurvey}, to keep the original topology of graphs, researchers begin to utilize the physical interaction between nodes in the graph to force the graph structures to be undamaged in the two or three dimensions~\cite{arleo2016distributed}. For instance, Haleem et al. \cite{haleem2019evaluating} combined a force-directed graph layout algorithm with deep learning algorithms to visualize the networks. The approach suggested has the advantage of  preserving the original graph structure by observing the physical attraction and repulsion between the nodes induced by force-directed graph layout algorithm. 

\begin{figure}[!t]
	
	\centering
	\includegraphics[scale=0.6]{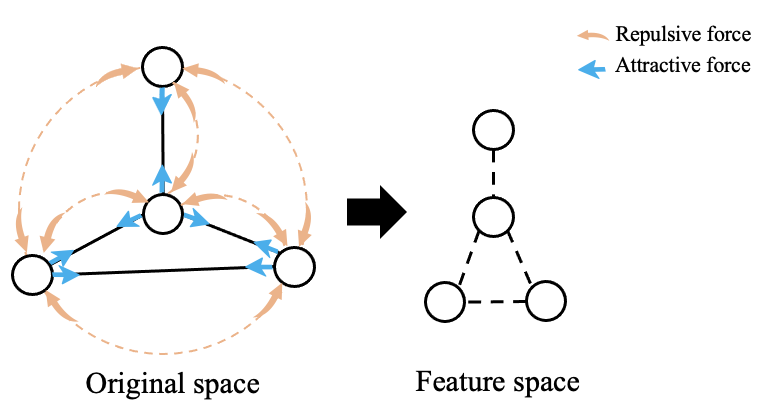}
	\caption{The example of graph representation considering physical properties. The force interaction of nodes (repulsive and attractive force) is presented to preserve the data structure in the feature space.}
	\label{2}
\end{figure}

By the application of physical principles, such as  repulsive and attractive force, on the layout of network structures, Sun et al~\cite{sun2020graph} proposed a GForce model. GForce is a graph learning model embedding the principles of force interaction related to physics. This helps in fully preserving the data structure in the vector space. Sun et al suggested that if the connected nodes have similarities, they are close to each other, otherwise, they are far from each other. However, in the feature space, the embedding vector of nodes may not represent the same structure as in the original graph. Thereby, inspired by the spring-electrical model \cite{ko2020flow}, they considered that there is a repulsive force between each node, and each node is attractive to its neighbours which are connected to it. In the feature space, due to the attraction, two connected nodes are relatively closer to each other, while unconnected nodes are far away from each other because of repulsion. Fig.~\ref{2} is an example of physics-based graph representation considering the force interaction of nodes. The graph structure can be well preserved in the feature space.

In the method suggested, the main step is importing attractive and repulsive relations, respectively. Given an undirected graph $G = {(V, E)}$, after embedding graph data into feature space, the feature vectors of the graph can be denoted as $\mathbf{H} \in  R^{|V|\times n}$. Here, the dimension of vector space is $n$ and satisfies $n\gg 3$. The feature vector $\mathbf{H}$ is updated by importing attractive and repulsive relations to the feature vectors:
\begin{equation}
	\mathbf{H} = \mathbf{H} + u \ast (\nabla F_a + \nabla F_r),
\end{equation}
here, $\nabla F_a $ and $\nabla F_r$ are the update of attractive and repulsive relations, respectively. $u$ is a parameter which is used to control the learning speed. The distance between nodes $i$ and $j$ in the vector space is $d_{ij} = \mathbf{h}_i - \mathbf{h}_j$, here $\mathbf{h}_i,\mathbf{h}_j \in \mathbf{H}$. The attractive relation of node $i$ to node $j$ is updated as the following:
\begin{equation}
	\nabla f_a(i) = -p \ast \widehat{w}_{ij}\ast d_{ij},
\end{equation}
where $p$ indicates the learning rate, and $\widehat{w}_{ij}\in \widehat{W}$ is the positive weight between nodes $i$ and $j$. On the other hand, the update of repulsive relation between node $i$ and node $j$ is as the following:

\begin{equation}
	\nabla f_r(i) = q \ast \widetilde{w}_{ij}\ast \frac{d_{ij}}{||d_{ij}|+ b|^2},
\end{equation}
where $q$ indicates the learning rate, $b$ denotes the distance bias, and $\widetilde{w}_{ij}\in \widetilde{W}$ is the negative weight between nodes $i$ and $j$. 

After importing the attractive relations, the connected nodes (or similar nodes) can be closer to each other in the feature vector space, while unconnected nodes (or different nodes) are far away from each other after importing repulsive relations. Sun et al~\cite{sun2020graph} evaluated their model on five datasets including Wikidata\footnote{https://www.wikidata.org/} Cora\footnote{https://relational.fit.cvut.cz/dataset/CORA}, Citeseer\footnote{https://relational.fit.cvut.cz/dataset/CiteSeer}, 20 Newsgroups\footnote{http://qwone.com/~jason/20Newsgroups/}, and WebKB\footnote{http://www.cs.cmu.edu/~webkb/}. 
Compared with various state-of-the-art baselines, such as SDNE~\cite{wang2016structural}, GForce showed better capacity in preserving the original structural information in low-dimensional feature spaces. It indicates that physics-inspired graph representation methods can effectively preserve the graph data structure in vector spaces.

\subsection{Physics-Driven Learning Models}

The step after representing training data in feature space, is model training. Specific learning algorithm needs to be identified to gain knowledge from training data and the trained model is then used in tasks such as classification and detection. There are still some challenges associated with the learning aspect of the models~\cite{9709096,9527112}. The challenges are for instance: loosing efficiency of the models by initialising using random parameters, model degradation by the imprecise loss function, lack of interpretation in deep graph learning models. The challenges are mitigated by physics-driven learning models proposed by researchers~\cite{alanazi2021survey}.

In this section, we introduce some advanced works focusing on physics-driven learning models. In particular, we  discuss how to reduce the overhead of models by initializing the parameters physically. Then, we give some examples of studies that improve the performance of learning models by adding physics laws to the loss functions.

\subsubsection{Parameter Initialization}

To improve the efficiency of training models and reduce the model overhead, recent works are carried out on informing the initial state of training models by satisfying some physical principles~\cite{shi2021physics,jia2021}. The main method is to use physical approaches to generate simulated data, which is subsequently used to pre-train the GL models ~\cite{schoenholz2020jax}. Therefore, the parameters of GL models can be reasonably initialized rather than getting randomly selected. The obvious advantage is that the problem associated with data paucity can be controlled in this way. 

Jia et al~\cite{jia2021physics} presented a physical technique to initialize recurrent graph network model, and then the pre-trained model is applied to predict the water temperature and flow in river networks. They first used a physical model to generate simulated target variables and intermediate physical variables. Then, the simulated variables were used to initialize and tune the GL model. The physical model they used is an energy balance model, which is based on the physics of thermodynamics and stream-flow. For example, the simulated temperature variables are generated according to the energy fluxes in the river segments. They took advantage of the relations between incoming and outgoing energy in the river segments to capture the temperature change. The temperature change conforms to the following equation:
\begin{equation}
	\Delta T \propto E_{in} - E_{out} + E_{up},
\end{equation}
where $\Delta T$ indicates the simulated temperature change, 
$E_{in}$ is the incoming energy fluxes, such as  rainfall and solar radiation, while $E_{out}$ indicates the outgoing energy fluxes, such as evaporation. $E_{up}$ is the net heat, which is advocated into the current river segment from the upstream segments. 

After using  simulated variables to pre-train the recurrent graph network model, they collected the real river segments data from the river network to train the model. Compared with other baselines, the model is more generalizable because the physical relations between river segments are enforced. Also, the pre-trained model will be more efficient when training is done with real data.

In addition, the simulated data can also help to alleviate the data paucity issues and reduce the data dependency of models. For instance, in the computer vision area, the simulated images have shown great ability to deal with real-world tasks including object localization. Shah et al.~\cite{shah2018airsim} proposed a method using a simulator, which is built on a video physical engine, that is used to pre-train the driving algorithm. Significantly, their method allows for less volume of training data.

\subsubsection{Loss Function}
\begin{figure*}[!t]
	
	\centering
	\subfigure{
		\begin{minipage}{1\linewidth}
			\centering
			\includegraphics[width= 3.5 in]{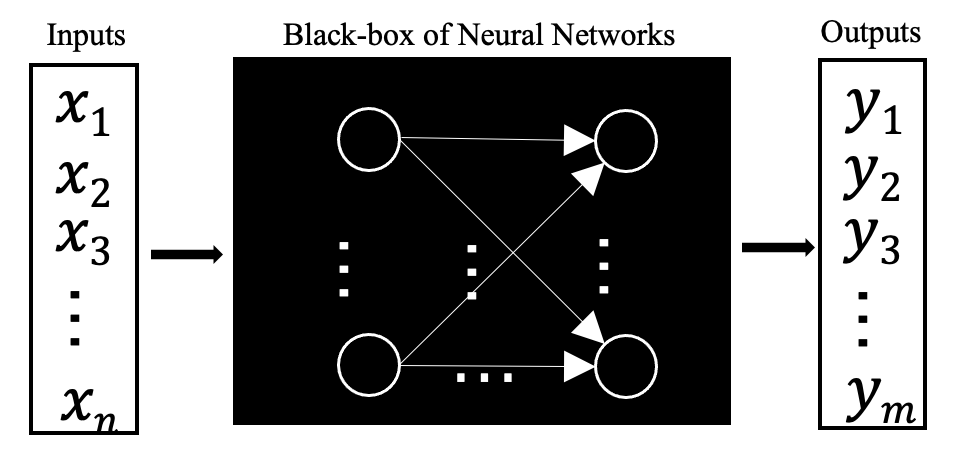} % 6.5cm
			
			(a) Traditional architecture of neural networks. %Due to the black-box property, nerual networks are unexplainable.
			
		\end{minipage}
	}
	\subfigure{
		\begin{minipage}{1\linewidth}
			\centering
			\includegraphics[width= 3.5 in]{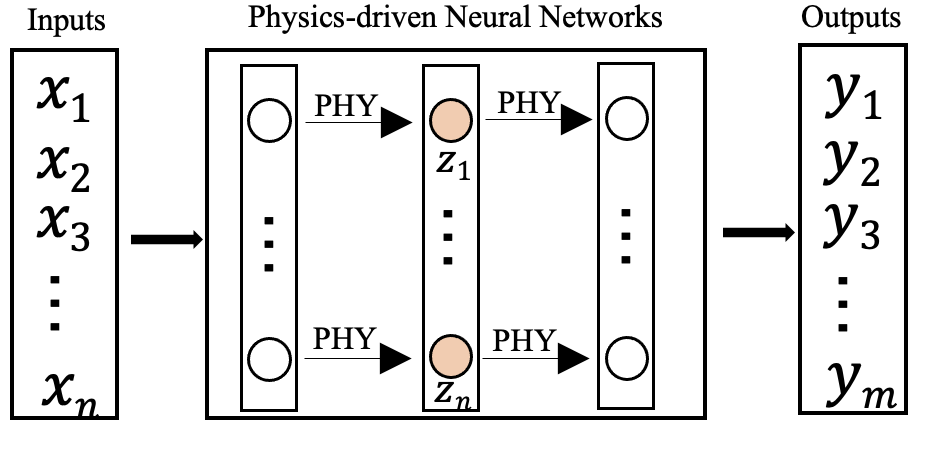}
			
			(b) Architecture of physics-driven neural networks.%Here, $z$ indicates the physical intermediate variables.
			
		\end{minipage}
	}	
	\caption{Comparison of traditional neural networks and physics-driven neural networks. As shown in (a), due to the black-box property, neural networks are unexplainable. In contrast, (b) is the neural networks infusing physics through physical intermediate variables $z$ or physics-guided connections among neurons. 
	}
	\label{3}
\end{figure*}

Data-driven models ignore the influence of physical variables and therefore, the standard GL model fails to gain advantage from the physical relationships of data. Numerous scientific problems emphasize the necessity to include physical constraints to explain the physical relationships of entities in data~\cite{li2022physics}. Therefore, to help GL models capture physical relationships and generalizable dynamic patterns which is consistent with known physical laws, researchers incorporate physical knowledge (e.g., physical intermediate variables) into the loss functions of GL models \cite{schuetz2021combinatorial}. The physical laws-informed loss function is expressed as follows:

\begin{equation}
	\begin{aligned}
		\mathcal{L} = \mathcal{L}_t(Y_{t},Y_{p}) + \lambda R(W) + \gamma \mathcal{L}_{p}(Y_{p}),
	\end{aligned}
\end{equation}
here, the training loss $\mathcal{L}_t$ measures errors between true values $Y_t$ and predicted values $Y_p$. The trade-off parameter $\lambda$ is to control the weight of regularization loss $R(W )$. $\mathcal{L}_t$ and $ \lambda R(W)$ are two terms of standard loss of learning models, while $\mathcal{L}_{p}$ denotes the physics-based loss. Particularly,  $\mathcal{L}_{p}$ is to ensure the predicted values $Y_p$ consistent with physical laws. $\gamma$ is the hyper-parameter weighted $\mathcal{L}_{p}$.

Daw et al~\cite{daw2020physics} presented a physics-guided loss function of the Long Short-Term Memory (LSTM) model to achieve accurate lake temperature prediction. Standard data-driven models typically only take the depth of water in lake data, that is, the position of nodes in graph data, as the main impact factor in temperature prediction. However, some other variables in lakes also have physical correlations with temperature. Among them, the temperature is closely related to density. That means temperature varies with density. Hence, in their method, not only the target temperature values $Y$ are considered, but also the physical intermediate values, density $Z$, are produced to build the loss function. They minimized the empirical loss over $Y$ and $Z$:

\begin{equation}
	\mathop{\arg\min} \mathcal{L}(Y_{t},Y_{p} )+ \lambda R(W) + \gamma \mathcal{L}_{p}(Z_{t},Z_{p}).
\end{equation}
Here, $Y_{t}$ and $Z_{t}$ denote the observed values of temperature and density, respectively.  $Y_{p}$ and $Z_{p}$ are output temperature and ancillary output density of the model.

Daw et al~\cite{daw2020physics} tested their model on two real lake datasets:  Lake Mendota and Falling Creek Reservoir. Their proposed model outperformed other LSTM-based methods when taking the root mean square error (RMSE) of models on the test datasets and physical inconsistency of lake data as the two evaluation metrics. 

As time or space changes, the observed data is affected by some physical factors. Therefore, compared with the general models, the models with the loss functions adding physical principles can have better performance. Moreover, physical laws provide reliable scientific theories to interpret neural networks~\cite{russell2022physics}. Thus, the explicable deep learning models can be significantly applied to scientific areas. Fig.~\ref{3} shows the comparison of traditional neural networks and physics-driven neural networks. 

\section{Open Challenges for PIGL}\label{OCFP}

Physics-informed graph learning expedited as more and more research works started incorporating physics models into graph learning models. Despite its obvious advantages, it still suffers from several challenges. This section briefly reviews some major challenges faced by PIGL.

\subsection{Cross-disciplinary Collaboration}

\noindent The integration of GL (AI and machine learning in general) and physics principles creates unprecedented potentials for addressing challenges that face GL, including data efficiency, trust and transparency. The realization of these potentials  demands effective interaction and collaboration among multiple disciplines. Most of current techniques for integrating GL and physics were developed by researchers from distinct disciplines and for isolated applications~\cite{park2019physics,ZHOU2021203}. To diminish this disciplinary gap, the pollination of research ideas and paradigms across diverse disciplines needs to be fostered.

\subsection{Physics-driven Co-design}

\noindent To alleviate specific problems such as data issues and model robustness, the existing PIGL models focus  on integrating physics into a certain step of GL. It might be worthwhile to embed interrelated physics at multiple steps. An example is to use cross-layer physical principles to deal with multiple problems~\cite{li2021physics}, for example, sparse data and data structure destruction. As such, the problems can be alleviated by taking advantage of relevant physics laws in multiple steps. 
In view of the advantages, an innovative PIGL design pattern/model that feature co-design of multiple (physics-based) modules may be devised. 

\subsection{Hybrid Systems}

Machine learning concepts are based on discrete time and space constraints. On the other hand, the physical concepts are realised in continuous time and space constraints~\cite{matsubara2020deep}. As the traditional graph learning approach is based on discreteness of the feature space, it is a tremendous challenge for the GL model to deal with tasks in real time constraints effectively. Therefore, by importing  the concept of continuous space-time into PIGL, the performance can be drastically improved. In other words, the PIGL can be regarded as a hybrid system, which combines both continuous and discrete systems ~\cite{li2021physics}. Due to the complex nature of hybrid systems, it is important to understand that, effective blending of continuous and discrete plays a major role in the performance of the PIGL models. An example could be demonstrating the changes in physical variables over time and space without breaking the properties of the data.

\subsection{Data Discrepancy}

The process of many PIGL methods is to first use physical models to generate synthetic data, and then train GL models by using the synthetic data. One obvious advantage associated with using synthetic data is that it can easily tackle data deficiency and missing problems  \cite{9780945}. In addition, synthetic data can be used to pre-train the GL models to initialize model parameters. However, due to the influence of various objective factors, the actual observation data could be inconsistent with the ideal state and so the ideal synthetic data might deviate from the actual observation data \cite{karniadakis2021physics}. Because of the discrepancy, the model trained with synthetic data will have errors when processing real tasks with real data. Therefore, one research pointer can be directed towards improving the robustness of PIGL models in the presence of discrepancy between the synthetic and actual data.

\subsection{Multi-domain Applications}
PIGL models have been applied to some specific tasks, such as power grid maintenance and water temperature detection. However, most of the current studies have  a limited range  of applications in scientific areas. Expanding the horizon of PIGL into diversified fields such as  finance and social networks~\cite{Zhang2020BGRdata} by taking full advantage of its potential is another promising research direction. An example could be, importing physical constraints into the social network data, and then using the processed data to train the GL model for completing assigned tasks such as relationship recognition in social networks~\cite{liu2021TKDE}.

\section{Conclusion}
Graph learning is one of the powerful AI techniques, attracting a great deal of attention from both academia and industry. Researchers have been working constantly on addressing challenges such as processing of complex graph data and solving the black-box property of deep learning. It could be a promising way to solve these problems by incorporating physics principles with GL. In principle, various physical laws can be effectively embedded into one or multiple steps of GL, yielding improved performance. To the best of our knowledge, this survey paper is the first review of PIGL. We expect that this survey paper will spark new interest in physics-informed design of graph learning systems. 

%\section*{Acknowledgment}

%The preferred spelling of the word ``acknowledgment'' in America is without 
%an ``e'' after the ``g''. Avoid the stilted expression ``one of us (R. B. 
%G.) thanks $\ldots$''. Instead, try ``R. B. G. thanks$\ldots$''. Put sponsor 
%acknowledgments in the unnumbered footnote on the first page.

\bibliographystyle{IEEEtran}
\bibliography{bibfile}

\end{document}